\documentclass[letterpaper, 10 pt, journal, twoside]{ieeetran}
\usepackage{amsmath,amsfonts}
\usepackage{array}
\usepackage[caption=false,font=normalsize,labelfont=sf,textfont=sf]{subfig}
\usepackage{textcomp}
\usepackage{stfloats}
\usepackage{url}
\usepackage{verbatim}
\usepackage{graphicx}
\usepackage{amssymb}

\usepackage{algorithm, algpseudocode}
\usepackage{tikz}
\usetikzlibrary{3d,arrows,positioning}

\usepackage{pgfplots}
\usepackage{booktabs}
\usepackage{graphicx, color, array, mathtools, cite, microtype, amsfonts, amsthm, amsmath, times, amssymb, multicol, multirow, siunitx}

\usepackage{subcaption}

\newtheorem{remark}{\bf{Remark}}

\def\BibTeX{{\rm B\kern-.05em{\sc i\kern-.025em b}\kern-.08em
    T\kern-.1667em\lower.7ex\hbox{E}\kern-.125emX}}
\usepackage{balance}
\begin{document}
\title{\LARGE GMP$^{3}$: Learning-Driven, Bellman-Guided Trajectory Planning for UAVs in Real-Time on SE(3)}
\author{Babak Salamat, Dominik Mattern, Sebastian-Sven Olzem, Gerhard Elsbacher, Christian Seidel, and Andrea M. Tonello%

\thanks{} 

}

\markboth{}
{Salamat \MakeLowercase{\textit{et al.}}: } 

\maketitle

\begin{abstract}
We propose \texttt{$\text{GMP}^{3}$}, a multiphase global path planning framework that generates dynamically feasible three-dimensional trajectories for unmanned aerial vehicles (UAVs) operating in cluttered environments.  The framework extends traditional path planning from Euclidean position spaces to the Lie group $\mathrm{SE}(3)$, allowing joint learning of translational motion and rotational dynamics. A modified Bellman-based operator is introduced to support reinforcement learning (RL) policy updates while leveraging prior trajectory information for improved convergence. \texttt{$\text{GMP}^{3}$} is designed as a distributed framework in which agents influence each other and share policy information along the trajectory: each agent refines its assigned segment and shares with its neighbors via a consensus-based scheme, enabling cooperative policy updates and convergence toward a path shaped globally even under kinematic constraints. We also propose \textit{DroneManager}, a modular ground control software that interfaces the planner with real UAV platforms via the MAVLink protocol, supporting real-time deployment and feedback. Simulation studies and indoor flight experiments validate the effectiveness of the proposed method in constrained 3D environments, demonstrating reliable obstacle avoidance and smooth, feasible trajectories across both position and orientation. The open-source implementation is available at \url{https://github.com/Domattee/DroneManager}.
\end{abstract}


\section{Introduction}\label{sec:1}
Recent advances in control systems have driven the widespread adoption of unmanned vehicles—including underwater, ground, and aerial systems (UAVs)—for a variety of applications. Classical approaches to motion planning, such as sampling-based methods, trajectory interpolation~\cite{9359893} or control based~\cite{9525383}, often rely on complete prior knowledge of the workspace and can struggle to scale or adapt to high dimensional or dynamic scenarios. As a result, many recent studies have focused on integrating machine learning and reinforcement leaning (RL) techniques into path planning to enable adaptability, improved performance, and broader applicability~\cite{UBBINK20209405, Farticle, 10473133}. 
\begin{figure}[t]
    \centering
    \begin{tikzpicture}[scale=1.3]
        \begin{axis}[
            view={15}{20}, 
            axis lines=center,
            xlabel={$x$}, ylabel={$y$}, zlabel={$z$},
            xlabel style={at={(xticklabel cs:1)}, anchor=west},
            ylabel style={at={(yticklabel cs:1.1)}, anchor=south},
            zlabel style={at={(axis cs:0,0,5.2)}, anchor=south},
            xtick=\empty, ytick=\empty, ztick=\empty,
            grid=major,
            xmin=0, xmax=5, ymin=0, ymax=5, zmin=0, zmax=5,
            enlargelimits=false
        ]
            \node at (5.2,0,0) {\small $x$};
            \node at (0,3.2,0) {\small $y$};
            \node at (0,0,3.2) {\small $z$};

           \addplot3[only marks, mark=*, color=green, mark options={draw=black, thin}] 
            coordinates {(0,0,0)}; 
            \node at (axis cs:0,0.5,0.3) {\footnotesize$S$};
            \addplot3[only marks, mark=*, color=green, mark options={draw=black, thin}] 
            coordinates {(4,2,3)}; 
            \node at (axis cs:4,2,3.5) {\footnotesize$G$};

            \addplot3[dashed, red, thick] coordinates {
                (0,0,0) (1,0.5,1) (2,1.5,1.2) (3.2,2,1.8) (4,2,3)
            } node[left] {\footnotesize$T'$};

            \addplot3[thin, blue] coordinates {
                (0,0,0) (2,0.05,0.5) (1.8,1.2,2) (3.8,1.7,1) (4,2,3)
            } node[right] {\footnotesize$T$};

            \addplot3[only marks, mark=*, color=black] coordinates {(2,0.05,0.5)} node[right] {\footnotesize$A_1$};
            \addplot3[only marks, mark=*, color=black] coordinates {(1.8,1.2,2)} node[above] {\footnotesize$A_2$};
            \addplot3[only marks, mark=*, color=black] coordinates {(3.8,1.7,1)} node[right] {\footnotesize$A_n$};

            \addplot3[only marks, mark=*, color=red!80] coordinates {(1,0.5,1)} node[below] {\footnotesize$A'_1$};
            \addplot3[only marks, mark=*, color=red!80] coordinates {(2,1.5,1.2)} node[below] {\footnotesize$A'_2$};
            \addplot3[only marks, mark=*, color=red!80] coordinates {(3.2,2,1.8)} node[above] {\footnotesize$A'_n$};

            \addplot3[->, thick, green] coordinates {(2,0.05,0.5) (1.1,0.5,1)} node[midway,right] {$\pi$};
            \addplot3[->, thick, green] coordinates {(1.8,1.2,2) (2,1.5,1.3)} node[midway,right] {$\pi$};
            \addplot3[->, thick, green] coordinates {(3.8,1.7,1) (3.2,2,1.7)} node[midway,right] {$\pi$};

            \addplot3[only marks, mark=*, mark size=10, color=blue!25, opacity=0.4] 
            coordinates {(0.5,0.5,1.8)} 
            node[anchor=center, xshift=0.0cm, yshift=0.0cm, text=black] {$\mathcal{O}_1$};
           \addplot3[only marks, mark=*, mark size=10, color=blue!25, opacity=0.4] 
            coordinates {(3,0.8,0.7)} 
            node[anchor=center, xshift=0.0cm, yshift=0.0cm, text=black] {$\mathcal{O}_2$};
            \addplot3[only marks, mark=*, mark size=10, color=blue!25, opacity=0.4] 
            coordinates {(2.8,1.5,2.8)} 
            node[anchor=center, xshift=0.0cm, yshift=0.0cm, text=black]{$\mathcal{O}_n$};
            \end{axis}
            \end{tikzpicture}
            \caption{3D visualization of the {\texttt{$\text{GMP}^{3}$}} framework, showing original ($T$) and perturbed ($T'$) trajectories, agent positions, and obstacles under policy $\pi$.}
            \label{GMP}
        \end{figure}
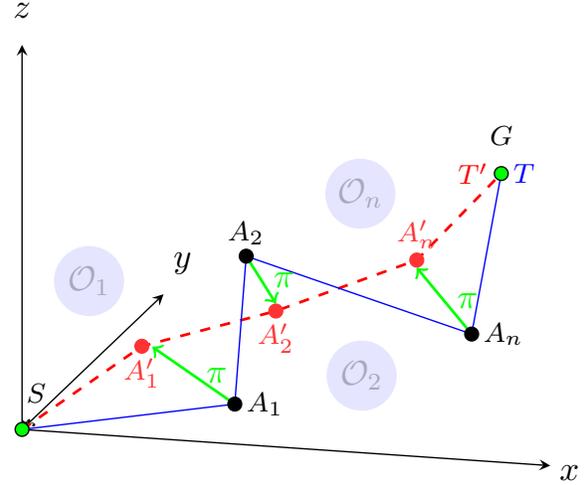
Active Simultaneous Localization and Mapping (SLAM) methods (e.g.,~\cite{10075065}) have explored how to acquire new information about an unknown environment, actively balancing exploration and exploitation. The relevance of these works brings attention to the necessity of policy design for navigation under limited or uncertain global knowledge. However, many active SLAM frameworks are not specifically designed for fully autonomous path planning in dynamic or cluttered spaces; they often focus on maintaining accurate localization and building maps, rather than enforcing strict dynamic feasibility or advanced multi-agent coordination. 

Deep neural networks have opened new avenues in path planning, as demonstrated in~\cite{9154607}, where Motion Planning Networks leverage a learned heuristic to generate sub-optimal paths. These architectures significantly reduce execution time compared to classical methods like RRT$^*$~\cite{5175292} or A$^*$~\cite{Astar}. On the other hand, they may require extensive offline training data and can suffer from domain shift when deployed in previously unseen contexts or dynamic environments. Similarly, cooperative motion planning in multi-robot or congested scenarios has been addressed through congestion-aware RL-based approaches~\cite{10829688}, which improves real-time decision-making yet often requires delicate reward tuning and can face challenges with global convergence under high robot density. In unknown and partially observable workspaces, an integral RL framework has been proposed to unify mapping, localization, and policy improvement under uncertain conditions~\cite{9785456}, but many solutions must still combat high sample complexity and may lack strong safety guarantees. For ground vehicles operating in harsh terrains, inverse RL has shown promise in learning sophisticated cost maps directly from expert trajectories~\cite{8629318}, although it can be sensitive to distribution mismatches between training and execution conditions. Other learning-based planners—such as those that combine global and local policies~\cite{9205217}, exploit dynamic graph representations for online adaptation~\cite{9205217}, or focus on UAV data gathering using time-varying Q-learning~\cite{10283892}—highlight the versatility of learning methods in a range of planning problems. Nevertheless, these approaches can be limited by high-dimensional action spaces, restrictive sensor assumptions, and the risk of suboptimal exploration strategies. However, developing and rigorously validating advanced control algorithms~\cite{9996578, 10171461}—and, by extension, inertial-navigation and sensor-fusion schemes~\cite{10054475, 9866842}—demands access to the full system state. This comprises not only the UAV’s translational motion (its position and the time evolution thereof) but also its rotational attitude, conventionally expressed by the pitch, roll, and yaw angles. To the best of the authors’ knowledge, no publicly available simulator can yet generate feasible representative full six–degree–of–freedom (6-DoF) trajectories that capture both translational and rotational dynamics with realistic characteristics, thereby limiting systematic controller design and benchmarking.

Recently, there has also been a surge of interest in adaptive RL-based UAV planners for search and rescue~\cite{10190142, 10316576} and in hybrid optimization learning schemes for accurate multi-joint manipulator motion~\cite{10097586}. While these methods show improved responsiveness and adaptability, they may encounter convergence hurdles or suffer from excessive computational costs when the planning horizon and action dimensionality grow. 
\subsection{Motivation and Contributions}
In recent advancements in reinforcement learning (RL) for trajectory optimization, the problem of autonomous path planning has been approached using Markov Decision Processes (MDPs) to minimize cumulative loss functions that account for trajectory smoothness, obstacle avoidance, and motion constraints. Traditional methods often rely on optimization-based approaches such as polynomial trajectory planning, which may not generalize well across dynamic environments. Instead, RL-based methods allow for adaptive learning of trajectory policies, where agents collaborate to modify and refine trajectories in real-time. Building on our previous work~\cite{10613437} that introduced the Global Multi-Phase Path Planning ({\texttt{$\text{GMP}^{3}$}}) concept in 2D environments, this paper significantly extends the framework and presents several new contributions:
\begin{itemize}
    \item Extension of the {\texttt{$\text{GMP}^{3}$}} framework to the Lie group $\mathrm{SE}(3)$, enabling full six-degree-of-freedom (6-DoF) trajectory planning that incorporates both translation and rotation.
    
    \item We introduce an influence-aware policy update rule that balances local adaptation, historical best performance, and global alignment.
    
    \item A consensus-based interaction protocol is employed to align neighboring agents through weighted policy sharing over a dynamic graph.
    
    \item  {\texttt{$\text{GMP}^{3}$}} offers a direct formulation and solution for policy optimization updates.
    
    \item Implementation of a modular, real-time drone control interface, \textit{DroneManager}, to deploy {\texttt{$\text{GMP}^{3}$}} on physical UAV platforms with minimal reconfiguration.
    
    \item Extensive validation in both simulation and real-world indoor experiments, demonstrating the robustness, adaptability, and practical viability of the proposed method in complex 3D environments.
\end{itemize}
Through a combination of global exploration, local refinement, and collaborative agent interaction, {\texttt{$\text{GMP}^{3}$}} can maintain high adaptability in dynamic or unknown environments while respecting kinematic and physical constraints. By integrating adaptive learning components and effective policy-sharing schemes, {\texttt{$\text{GMP}^{3}$}} aspires to outperform existing single-stage or solely local solutions in terms of scalability, convergence speed, and planning quality. This approach facilitates the creation of rapid, feasible trajectories under real-time constraints—an essential capability for drones, ground robots, and other platforms operating in cluttered or safety-critical domains. In the remainder of this paper, we present the details of our {\texttt{$\text{GMP}^{3}$}} algorithm, discuss its connection to existing RL-based methods, and validate its performance in both simulation and hardware experiments. The results highlight the benefits of leveraging a multi-phase perspective for robust and efficient path planning in challenging scenarios.

The structure of the paper is as follows. Section~\ref{sec:2} defines the formal problem and introduces the multi-phase trajectory learning setup in a three-dimensional environment. Section~\ref{sec:3} presents the reinforcement learning-based update mechanism, including the formulation of a modified Bellman operator tailored to the trajectory optimization framework. Section~\ref{sec:4} describes five optimization strategies employed to enhance convergence speed and learning stability. Section~\ref{sec:5} reports simulation results that evaluate the effectiveness of the proposed method under various scenarios. Section~\ref{sec:6} introduces the custom-developed DroneManager software, which facilitates real-time control and coordination of physical UAVs. Section~\ref{sec:7} presents the results of real-world experiments that validate the practicality and robustness of the proposed approach. The paper concludes with a summary of findings and outlines potential directions for future research.

\textit{ Notations:} The notation $\langle X \rangle$ denotes the mean of the quantity $X$.

\section{Problem Statement}\label{sec:2}
We consider the challenge of navigating a rigid body, such as an unmanned aerial vehicle (UAV), through a cluttered 3D environment. The system dynamics are approximated as kinematic constraints, with bounded position, velocity, and acceleration. The primary objective is to develop a multi-phase, learning-based framework to generate and refine feasible trajectories in real-time. We aim to answer: (i) What features should be learned for efficient trajectory shaping? (ii) How can a distributed learning framework be designed to enable effective policy updates using influence-aware policy update mechanisms?
\subsection{Multi-Phase Trajectory Learning Setup in 3D Space}
We propose a global multi-phase path planning framework, \texttt{GMP$^3$}, extended to 3D~(see Fig.~\ref{GMP}). Let $n$ denote the number of agents in a workspace $\mathbb{R}^3$ with known static obstacles. Each agent $A_i$ ($i=1,\ldots,n$) collaborates to shape the global trajectory. The system is formulated as a Markov Decision Process (MDP):
\begin{equation}
\mathcal{M} = (\mathcal{S}, \mathcal{A}, \mathcal{P}, \mathcal{R}, \gamma),
\end{equation}
where the state space and action space are defined as follows:
\begin{align}
\mathcal{S} &\coloneqq \bigcup_{i=1}^n \mathcal{S}_i \\
\mathcal{A} &\coloneqq \bigcup_{i=1}^n \mathcal{A}_i
\end{align}
Each agent $A_i$ at time $t$ is represented by a position vector $\mathbf{p}_{t,i} \in \mathbb{R}^3$. The start and goal positions are represented by vectors $\mathbf{p}_s, \mathbf{p}_g \in \mathbb{R}^3$, respectively. The obstacle set is defined as $\mathcal{O} = \{(\mathbf{o}_k, r^{k}_{\text{obs}}) \mid \mathbf{o}_k \in \mathbb{R}^3, r^{k}_{\text{obs}} > 0, \ k = 1,\dots,n_{\text{obs}}\}$. The full system state at time $t$ is given by:
\begin{equation}
\mathcal{S}_t = \{ \mathbf{p}_{t,i} \}_{i=1}^n \cup \{ \mathbf{p}_s, \mathbf{p}_g \} \cup \mathcal{O}.
\end{equation}
The action for agent $A_i$ at time $t$ is a perturbation vector $\boldsymbol{\delta}_{t,i} \in \mathbb{R}^3$, and the joint action is defined as:
\begin{equation}
\mathcal{A}_t = \{ \boldsymbol{\delta}_{t,i} \}_{i=1}^n.
\end{equation}
The obstacle set is defined as $\mathcal{O} = \{(o_k, r^{k}_{\text{obs}}) \mid o_k \in \mathbb{R}^3, r^{k}_{\text{obs}} \in \mathbb{R}_{>0}, 1 \leq k \leq n_{\text{obs}}\}$. The state transition follows deterministic kinematics: $\mathcal{S}_{t+1} = \mathcal{S}_t + \mathcal{A}_t$, and the reward function is derived from a loss function $\mathcal{L}(\mathcal{S}, \mathcal{A})$ such that $\mathcal{R} = -\mathcal{L}$. The scalar $\gamma$ is a discount factor in the interval $[0,1]$.
\subsection{State and Action Representation}
Each position vector $\mathbf{p}_{t,i}$ associated with agent $A_i$ at time step $t$ is an element of $\mathbb{R}^3$, and is explicitly defined as $\mathbf{p}_{t,i} = (x_{t,i}, y_{t,i}, z_{t,i})^\top$. This vector represents the spatial coordinates of the agent in the three-dimensional workspace, encapsulating its position along the $x$, $y$, and $z$ axes. Similarly, the action vector $\boldsymbol{\delta}_{t,i}$, which governs the motion of agent $A_i$ at time $t$, is also an element of $\mathbb{R}^3$, given by $\boldsymbol{\delta}_{t,i} = (\Delta x_{t,i}, \Delta y_{t,i}, \Delta z_{t,i})^\top$. Each component of this vector corresponds to a perturbation or control adjustment applied in the respective spatial direction. The evolution of the system is governed by a deterministic transition model. Specifically, the position of each agent is updated based on the action it executes at the current time step. The update rule is described by the following equation:
\begin{equation}
\mathbf{p}_{t+1,i} = \mathbf{p}_{t,i} + \boldsymbol{\delta}_{t,i}, \quad \forall i = 1, \dots, n.
\end{equation}
This transition model implies that the position at the next time step is obtained by adding the perturbation vector to the current position vector.  The update is applied synchronously across all agents, which supports coordinated trajectory shaping in the multi-phase planning framework.
\subsection{Rigid–Body State on {$\mathrm{SE}(3)$}}
To enable the planner with full 6-DoF capability, we shift the state of every agent from Euclidean positions to rigid–body poses on the Lie group
\begin{align}
\mathrm{SE}(3)\;=\;\Bigl\{
\begin{bsmallmatrix}
\mathbf{R} & \mathbf{p} \\ \mathbf{0}_{1\times 3} & 1
\end{bsmallmatrix}
\bigm|
\mathbf{R}\in\mathrm{SO}(3),\;
\mathbf{p}\in\mathbb{R}^{3}
\Bigr\}.
\end{align}
At time~$t$, agent~$A_i$ is described by  
\begin{align}
\mathbf{T}_{t,i}\;=\;
\begin{bmatrix}
\mathbf{R}_{t,i} & \mathbf{p}_{t,i}\\
\mathbf{0}_{1\times 3} & 1
\end{bmatrix}\in\mathrm{SE}(3),
\end{align}
where $\mathbf{R}_{t,i}\!\in\!\mathrm{SO}(3)$ is the orientation matrix and
$\mathbf{p}_{t,i}\!\in\!\mathbb{R}^{3}$ the position used in the previous
subsections.  The collection of all agent poses, together with
$\{\mathbf{p}_s,\mathbf{p}_g\}$ and the obstacle set~$\mathcal{O}$, forms the
extended state $\mathcal{S}_t$. An action now lives in the Lie algebra $\mathfrak{se}(3)$ and is expressed by
the twist
\begin{align}
\boldsymbol{\xi}_{t,i}
=\begin{bmatrix}\boldsymbol{v}_{t,i}\\ \boldsymbol{\omega}_{t,i}\end{bmatrix}
\in\mathbb{R}^{6},\qquad
\boldsymbol{v}_{t,i},\boldsymbol{\omega}_{t,i}\in\mathbb{R}^{3},
\end{align}
with linear and angular components, respectively. Using the \emph{hat} operator
\begin{align}
\widehat{\boldsymbol{\xi}}
=\begin{bmatrix}
[\boldsymbol{\omega}]_\times & \boldsymbol{v}\\
\mathbf{0}_{1\times 3} & 0
\end{bmatrix},\qquad
[\boldsymbol{\omega}]_\times=\begin{bmatrix}
0&-\omega_3&\ \omega_2\\
\omega_3&0&-\omega_1\\
-\omega_2&\ \omega_1&0
\end{bmatrix},
\end{align}
the deterministic kinematics adopt the invariant form
\begin{align}\label{SEt}
\mathbf{T}_{t+1,i}
=\mathbf{T}_{t,i}\,
\exp\!\bigl(\! \Delta t\,\widehat{\boldsymbol{\xi}_{t,i}}\bigr)
,\quad\forall i=1,\dots,n,
\end{align}
where $\Delta t\equiv dt$ is the time–step already used in
Table~\ref{app:gmpparams}, and $\exp$ is the matrix exponential on $\mathfrak{se}(3)\rightarrow\mathrm{SE}(3)$.
For a discrete‐time trajectory
$\{\mathbf{T}_{t_j,i}\}_{j=0}^{N}$ with increments
$\Delta\mathbf{p}_{t_j,i}=\mathbf{p}_{t_{j+1},i}-\mathbf{p}_{t_j,i}$, define
the incremental rotational distance
\begin{align}
d_R\!\bigl(\mathbf{R}_{t_j,i},\mathbf{R}_{t_{j+1},i}\bigr)
=\frac{1}{\sqrt{2}}\,
\bigl\|\log\!\bigl(\mathbf{R}_{t_j,i}^\top\mathbf{R}_{t_{j+1},i}\bigr)\bigr\|_F,
\end{align}
where $\log:\mathrm{SO}(3)\!\rightarrow\!\mathfrak{so}(3)$ is the matrix logarithm and $\|\cdot\|_F$ the Frobenius norm.
\section{Reinforcement Learning-Based Trajectory Update Rule}\label{sec:3}
We cast the trajectory refinement process in the \texttt{GMP$^{3}$} framework as a distributed reinforcement learning (RL) problem. Each agent $A_i$ learns a policy that produces body-frame twists $\boldsymbol{\xi}_{t,i} \in \mathbb{R}^6$ to minimize a global $\mathrm{SE}(3)$-aware loss while coordinating with neighboring agents under a consensus protocol.
Let $\pi_i : \mathcal{S} \to \mathbb{R}^6$ denote the local policy of agent $A_i$, which maps the extended state $\mathcal{S}_t \in \mathbb{R}^{6n + 6 + 4n_{\text{obs}}}$ to a twist vector $\boldsymbol{\xi}_{t,i} \in \mathbb{R}^6$, consisting of linear and angular velocity components. The joint policy at time $t$ is defined as:
\begin{equation}
\pi_t(\mathcal{S}_t) = \left\{ \pi_i(\mathcal{S}_t) \right\}_{i=1}^n = \mathcal{A}_t.
\end{equation}
Each agent aims to minimize a global cost defined by an $\mathrm{SE}(3)$-aware loss function $\mathcal{L}_{\mathrm{SE(3)}}(\mathcal{S}_t, \mathcal{A}_t)$, which penalizes deviations from smooth motion, safe obstacle clearance, and rotational inconsistency. The learning objective is:
\begin{equation}
\min_{\pi} \quad \mathbb{E}_{\mathcal{S}_t \sim \rho^\pi} \left[ \mathcal{L}_{\mathrm{SE(3)}}(\mathcal{S}_t, \pi(\mathcal{S}_t)) \right],
\end{equation}
where $\rho^\pi$ denotes the distribution over trajectories induced by the policy $\pi$ and the deterministic$\mathrm{SE}(3)$ dynamics.
\subsection{Influence-Aware Policy Update}
To ensure coordinated learning along the shared trajectory, we introduce an influence-aware policy update rule that combines elements of personal adaptation and alignment with globally effective behaviors. Let $\pi_i : \mathcal{S} \rightarrow \mathbb{R}^6$ denote the local policy of agent $A_i$, mapping the extended system state $\mathcal{S}_t$ to a 6D twist $\boldsymbol{\xi}_{t,i} = (\boldsymbol{v}_{t,i}^\top, \boldsymbol{\omega}_{t,i}^\top)^\top$, where $\boldsymbol{v}_{t,i}, \boldsymbol{\omega}_{t,i} \in \mathbb{R}^3$ represent linear and angular velocities, respectively. The joint action at time $t$ is then:
\begin{align}
    \mathcal{A}_t = \{\boldsymbol{\xi}_{t,i}\}_{i=1}^{n} \in (\mathbb{R}^6)^n.
\end{align}
Each agent maintains two memory-based policies: a personal best policy $\pi_i^l$, which minimizes its own historical trajectory cost, and a global best policy $\pi^g$, which corresponds to the lowest cost observed across all agents in the network. The influence-aware policy update rule combines local loss descent with attraction toward both personal and global optima:
\begin{align}
\pi_{t+1,i} = \pi_{t,i} 
&- \alpha \nabla_{\pi_i} \mathcal{L}_{\mathrm{SE(3)}}(\mathcal{S}_t, \pi_t(\mathcal{S}_t)) \notag \\
&- \beta_1 (\pi_{t,i} - \pi^l_i) - \beta_2 (\pi_{t,i} - \pi^g),
\end{align}
where $\alpha > 0$ is the learning rate, and $\beta_1, \beta_2 > 0$ are influence coefficients balancing personal adaptation and global convergence.
\subsection{Mutual Alignment Through Local Influence}
To enforce agreement among neighboring agents, a mutual alignment term is introduced. Let $\mathcal{N}_i$ denote the neighbor set of agent $A_i$, and $w_{ij} \geq 0$ be the consensus weights with $\sum_{j \in \mathcal{N}_i} w_{ij} = 1$.
The full policy update, incorporating mutual alignment with neighbors, becomes:
\begin{align}
\pi_{t+1,i} = \pi_{t,i} 
&- \alpha \nabla_{\pi_i} \mathcal{L}_{\mathrm{SE(3)}}(\mathcal{S}_t, \pi_t(\mathcal{S}_t)) \notag \\
&- \beta_1 (\pi_{t,i} - \pi^l_i) - \beta_2 (\pi_{t,i} - \pi^g) \notag \\
&- \sum_{j \in \mathcal{N}_i} w_{ij} (\pi_{t,i} - \pi_{t,j}). \label{eq:se3_update}
\end{align}
This consensus-based structure allows agents to refine their policy based on their own feedback while staying close to the behavior of neighboring agents along the multi-phase trajectory.
\subsection{Policy Evaluation and Improvement}
The expected performance of a policy $\pi$ is quantified via the value function:
\begin{align}
\mathcal{V}^\pi(\mathcal{S}) = \mathbb{E}_\pi \left[ \sum_{k=0}^{\infty} \gamma^k \left( -\mathcal{L}_{\mathrm{SE(3)}}(\mathcal{S}_{t+k}, \pi(\mathcal{S}_{t+k})) \right) \mid \mathcal{S}_t = \mathcal{S} \right],
\end{align}
where $\gamma \in [0,1]$ is a discount factor and $\mathcal{L}_{\mathrm{SE(3)}}$ is the $\mathrm{SE}(3)$-aware loss defined later in this section.
Since analytic gradients of $\mathcal{L}_{\mathrm{SE(3)}}$ with respect to the twist $\boldsymbol{\xi}_{t,i}$ are typically unavailable, we use finite-difference approximations. These approximations, presented in Table~\ref{tab:policy_formulas}, include the two-point, three-point, and five-point formulas, each offering increasingly higher-order accuracy.
\begin{table*}[ht]
\centering
\caption{Finite-Difference Approximations for $\mathrm{SE}(3)$ Policy Gradient}
\label{tab:policy_formulas}
\begin{tabular}{@{}p{0.97\linewidth}@{}}
\toprule
\textbf{Two-Point} \\
\[
\pi(\mathcal{S}_t) \approx 
\frac{
\mathcal{L}_{\mathrm{SE(3)}}(\mathcal{S}_t, \mathcal{A}_t + \Delta \mathbf{e}_k)
-
\mathcal{L}_{\mathrm{SE(3)}}(\mathcal{S}_t, \mathcal{A}_t - \Delta \mathbf{e}_k)
}{2 \Delta}
\]
\\[1ex]
\textbf{Three-Point} \\
\[
\pi(\mathcal{S}_t) \approx 
\frac{
4\mathcal{L}_{\mathrm{SE(3)}}(\mathcal{S}_t, \mathcal{A}_t + \Delta \mathbf{e}_k)
- \mathcal{L}_{\mathrm{SE(3)}}(\mathcal{S}_t, \mathcal{A}_t + 2\Delta \mathbf{e}_k)
- 3\mathcal{L}_{\mathrm{SE(3)}}(\mathcal{S}_t, \mathcal{A}_t)
}{2 \Delta}
\]
\\[1ex]
\textbf{Five-Point} \\
\[
\pi(\mathcal{S}_t) \approx 
\frac{
-\mathcal{L}_{\mathrm{SE(3)}}(\mathcal{S}_t, \mathcal{A}_t + 2\Delta \mathbf{e}_k)
+ 8\mathcal{L}_{\mathrm{SE(3)}}(\mathcal{S}_t, \mathcal{A}_t + \Delta \mathbf{e}_k)
- 8\mathcal{L}_{\mathrm{SE(3)}}(\mathcal{S}_t, \mathcal{A}_t - \Delta \mathbf{e}_k)
+ \mathcal{L}_{\mathrm{SE(3)}}(\mathcal{S}_t, \mathcal{A}_t - 2\Delta \mathbf{e}_k)
}{12 \Delta}
\]
\\
\midrule
\textit{Note:} $\xi_{(k),i}$ denotes the $k$-th component of the twist $\boldsymbol{\xi}_{t,i} \in \mathbb{R}^6$ for agent $A_i$, and $\mathbf{e}_k$ is the corresponding standard basis vector in $\mathbb{R}^{6n}$. The scalar $\Delta$ is a small perturbation magnitude used to approximate the directional derivative of the loss function $\mathcal{L}_{\mathrm{SE(3)}}$. \\
\bottomrule
\end{tabular}
\end{table*}
\subsection{Bellman Operator for {\texttt{$\text{GMP}^{3}$}}}
To formalize distributed reinforcement learning with consensus, we define a modified Bellman operator that incorporates the structured communication policy in \texttt{GMP$^3$}. The optimal policy satisfies a Bellman recursion adapted to the consensus-aware $\mathrm{SE}(3)$ planning problem. The distributed Bellman operator $\mathcal{T}$ is defined as:
\begin{align}
(\mathcal{T} V)(\mathcal{S}) = \min_{\{\pi_i\}} \left[
\mathcal{L}_{\mathrm{SE(3)}}\left( \mathcal{S}, \{ \pi_i(\mathcal{S}) \}_{i=1}^n \right)
+ \gamma \, V(\mathcal{S}') \right],
\end{align}
where $\mathcal{S}'$ is the deterministic next state obtained via SE(3) transitions in~\eqref{SEt}. 
\subsection{Loss Function and Obstacle Violation Term}
We formalize the policy–learning objective by introducing a trajectory-based
loss that couples translational smoothness, rotational smoothness, and
proximity to obstacles. Let
$t_0 < t_1 < \dots < t_N$ denote $N + 1$ fixed sampling instants, and
let $\mathbf{T}_{t_j} =
\begin{bmatrix}
\mathbf{R}_{t_j} & \mathbf{p}_{t_j} \\
\mathbf{0}_{1\times3} & 1
\end{bmatrix} \in \mathrm{SE}(3)$
denote the rigid–body pose at time $t_j$. The translational increment between time steps is defined as
\[
\Delta\mathbf{p}_{t_j} = \mathbf{p}_{t_{j+1}} - \mathbf{p}_{t_j}, \qquad j = 0,\dots,N-1,
\]
and the rotational displacement is measured via the geodesic distance on $\mathrm{SO}(3)$:
\[
d_R(\mathbf{R}_{t_j}, \mathbf{R}_{t_{j+1}}) = \frac{1}{\sqrt{2}} \left\| \log\left( \mathbf{R}_{t_j}^\top \mathbf{R}_{t_{j+1}} \right) \right\|_F,
\]
where $\log(\cdot)$ denotes the matrix logarithm and $\|\cdot\|_F$ the Frobenius norm.

Let $Q \in \mathbb{R}^{3 \times 3}$ be a symmetric positive definite matrix that penalizes translational non-smoothness by weighting velocity increments. 

To account for obstacle proximity, we incorporate a violation penalty. For each obstacle \( k = 1, \dots, n_{\text{obs}} \), let \( \mathbf{o}_k \in \mathbb{R}^3 \) denote the obstacle center and \( r^k_{\text{obs}} > 0 \) its influence radius. The Euclidean distance from the pose at time \( t_j \) to obstacle \( k \) is given by \( d_k^{(j)} = \lVert \mathbf{p}_{t_j} - \mathbf{o}_k \rVert_2 \). The obstacle violation term is defined as the average proximity penalty across all trajectory points:
\begin{equation} \label{violation}
\nu = \sum_{k=1}^{n_{\text{obs}}} \frac{1}{N+1} \sum_{j=0}^N \max\left(1 - \frac{d_k^{(j)}}{r^k_{\text{obs}}}, 0\right).
\end{equation}
This penalty increases as the agent approaches or enters the obstacle's influence region.

The total $\mathrm{SE}(3)$-aware loss function that governs the reinforcement learning process is then given by:
\begin{align}
\mathcal{L}_{\mathrm{SE(3)}}(\mathcal{S}, \mathcal{A}) 
&= \sum_{j=0}^{N-1} \left[
(\Delta \mathbf{p}_{t_j})^\top Q \Delta \mathbf{p}_{t_j} \right. \notag \\
&\left. \quad + \mu \cdot d_R^2(\mathbf{R}_{t_j}, \mathbf{R}_{t_{j+1}})
\right] \cdot (1 + \lambda \nu),
\end{align}
where \( \lambda > 0 \) is a weighting parameter that trades off between trajectory regularity and obstacle avoidance. To similarly regularize orientation transitions, we introduce a scalar coefficient $\mu > 0$ that penalizes squared rotational deviations. 
\begin{algorithm}
\caption{RL Components of \texttt{GMP$^3$} ($\mathrm{SE}(3)$-aware)}
\begin{algorithmic}[0]
  \Require $\mathrm{SE}(3)$-aware loss function $\mathcal{L}_{\mathrm{SE(3)}}(\mathcal{S})$
  \State Initialize state $\mathcal{S}_{t=0}$, policy $\pi_{0,i}$, value $\mathcal{V}_{0,i}$ for all $i = 1, \dots, n$
  \For{$t = 0, 1, \dots, T$}
    \For{$i = 1, 2, \dots, n$}
      \[
      \big\{\Delta_{\xi_{(1),i}}, \dots, \Delta_{\xi_{(6),i}}\big\}
      \leftarrow \nabla_{\pi_i} \mathcal{L}_{\mathrm{SE(3)}}(\mathcal{S}_t)
      \]
      \If{stopping criterion is met}
        \State \Return $\mathcal{V}^{\pi^\ast}(\mathcal{S})$
      \EndIf
      \[
      \mathcal{V}_{t+1,i} = \mathcal{V}_{t,i} - \nabla_{\pi_i} \mathcal{L}_{\mathrm{SE(3)}}(\mathcal{S}_t)
      \]
    \EndFor
  \EndFor
\end{algorithmic}
\label{Algo_1}
\end{algorithm}
\section{Optimization Strategies in \texttt{$\text{GMP}^{3}$}} \label{sec:4}
This section reviews the optimisation methods used in
\texttt{GMP$^{3}$} to update the policies
$\{\pi_i\}_{i=1}^{n}$ by minimising the
consensus–regularised, obstacle–aware loss
$\mathcal{L}_{\mathrm{SE(3)}}(\mathcal{S},\mathcal{A})$
introduced in Section~\ref{sec:3}.  The value–function update in
Algorithm~\ref{Algo_1} is interpreted as a gradient‐based optimisation scheme, and can be combined with several modern stochastic optimisers including Momentum Gradient Descent (MGD)~\cite{MGD}, AdaGrad~\cite{duchi2011adaptive}, RMSProp~\cite{tieleman2012rmsprop}, AdaDelta~\cite{AdaDelta}, and Adam~\cite{Adam}.
Let $\boldsymbol{\xi}_{t,i}=\pi_{t,i}(\mathcal{S}_t)\in\mathbb{R}^{6}$
be the twist returned by the current policy of agent~$A_i$.
Define the composite update direction
\begin{align}\nonumber
\Pi :=\;
&\alpha \, \nabla_{\pi_i}
        \mathcal{L}_{\mathrm{SE(3)}}\!\bigl(\mathcal{S}_t,\pi_t(\mathcal{S}_t)\bigr)
\;+\;
\beta_1 \bigl(\pi_{t,i}-\pi^{l}_i\bigr) \\
\;+\; 
&\beta_2 \bigl(\pi_{t,i}-\pi^{g}\bigr) \label{PU}
\;+\;
\sum_{j\in\mathcal{N}_i} w_{ij}\bigl(\pi_{t,i}-\pi_{t,j}\bigr),
\end{align}
where $\alpha$ is the learning rate,
$\beta_{1,2}>0$ weight the influence–aware terms,
and $\{w_{ij}\}$ are the consensus weights on the neighbour set
$\mathcal{N}_i$.  A plain gradient descent step reads
\begin{equation}
\pi_{t+1,i} \;=\; \pi_{t,i} \;-\; \Pi.
\end{equation}
This operation is embedded within the broader consensus-leader dynamics described previously, and the optimization aims to strike a balance between local policy improvements and global alignment. In the following, we describe different optimization techniques that refine this update process.
\begin{figure*}[]
    \centering
    \includegraphics[width=\linewidth]{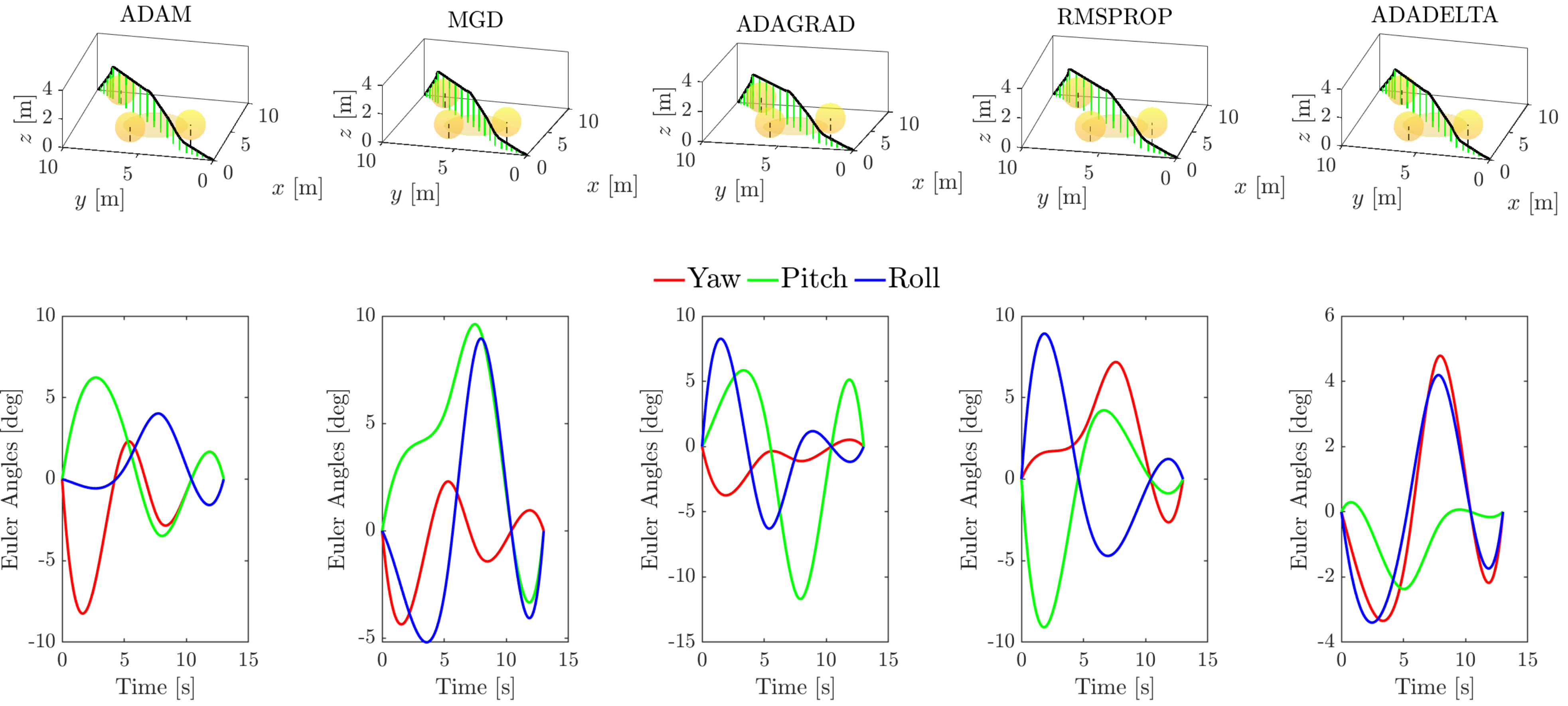}
    \caption{$\mathrm{SE}(3)$-aware {\texttt{$\text{GMP}^{3}$}} trajectories and Euler angles over time. Top: 3D position trajectories overlaid with altitude stalks and obstacles. Bottom: Euler angle evolution (yaw: red, pitch: green, roll: blue).}
    \label{fig:3D_ex}
\end{figure*}
\subsection{Momentum Gradient Descent (MGD)}
Momentum Gradient Descent (MGD) accelerates convergence by incorporating a velocity term that accumulates the past gradients.
\begin{align}
    v_{t+1} &= \beta v_t - \Pi, \\
    \pi_{t+1} &= \pi_t + v_{t+1}
\end{align}
\begin{algorithm}[H]
\caption{Momentum Gradient Descent for \texttt{$\text{GMP}^{3}$}}
\begin{algorithmic}[0]
    \Require Learning rates $\alpha, \beta_1, \beta_2$, momentum coefficient $\beta$
    \State Initialize policy $\pi_0$, velocity $v_0 \gets 0$
    \For{$t = 1$ to $n$}
        \State $g_t \gets  \ \Pi$
        \State $v_t \gets \beta v_{t-1} - g_t$
        \State $\pi_t \gets \pi_{t-1} + v_t$
    \EndFor
\end{algorithmic}
\end{algorithm}
\subsection{Adaptive Gradient Algorithm (AdaGrad)}
AdaGrad adapts the learning rate for each parameter by scaling it inversely proportional to the square root of the sum of all historical squared gradients.
\begin{align}
    G_t &= G_{t-1} +  \Pi^2, \\
    \pi_{t+1} &= \pi_t - \frac{\eta}{\sqrt{G_t + \epsilon}}  \ \Pi
\end{align}
\begin{algorithm}[H]
\caption{AdaGrad for \texttt{$\text{GMP}^{3}$}}
\begin{algorithmic}[0]
    \Require Learning rates $\alpha, \beta_1, \beta_2$, small constant $\epsilon$
    \State Initialize policy $\pi_0$, $G_0 \gets 0$
    \For{$t = 1$ to $n$}
        \State $g_t \gets  \ \Pi$
        \State $G_t \gets G_{t-1} +  \ \Pi^2$
        \State $\pi_t \gets \pi_{t-1} - \frac{1}{\sqrt{G_t + \epsilon}}  \ \Pi$
    \EndFor
\end{algorithmic}
\end{algorithm}
\subsection{Root Mean Square Propagation (RMSProp)}
RMSProp modifies AdaGrad by using an exponentially decaying average to prevent the learning rate from diminishing too quickly.
\begin{align}
    E[g^2]_t &= \gamma E[g^2]_{t-1} + (1 - \gamma)  \ \Pi^2, \\
    \pi_{t+1} &= \pi_t - \frac{\eta}{\sqrt{E[g^2]_t + \epsilon}}  \ \Pi
\end{align}
\begin{algorithm}[H]
\caption{RMSProp for \texttt{$\text{GMP}^{3}$}}
\begin{algorithmic}[0]
    \Require Learning rates $\alpha, \beta_1, \beta_2$, decay factor $\gamma$, small constant $\epsilon$
    \State Initialize policy $\pi_0$, $E[g^2]_0 \gets 0$
    \For{$t = 1$ to $n$}
        \State $g_t \gets  \ \Pi$
        \State $E[g^2]_t \gets \gamma E[g^2]_{t-1} + (1 - \gamma) g_t^2$
        \State $\pi_t \gets \pi_{t-1} - \frac{1}{\sqrt{E[g^2]_t + \epsilon}} g_t$
    \EndFor
\end{algorithmic}
\end{algorithm}
\subsection{Adaptive Delta (AdaDelta)}
AdaDelta adjusts the learning rate based on the ratio of exponentially weighted averages of past gradients and updates.
\begin{align}
    E[g^2]_t &= \rho E[g^2]_{t-1} + (1 - \rho)  \ \Pi^2, \\
    \Delta \pi_t &= - \frac{\sqrt{E[\Delta \pi^2]_{t-1} + \epsilon}}{\sqrt{E[g^2]_t + \epsilon}}  \ \Pi, \\
    \pi_{t+1} &= \pi_t + \Delta \pi_t
\end{align}
\begin{algorithm}[H]
\caption{AdaDelta for \texttt{$\text{GMP}^{3}$}}
\begin{algorithmic}[0]
    \Require Decay factor $\rho$, small constant $\epsilon$
    \State Initialize policy $\pi_0$, $E[g^2]_0 \gets 0$, $E[\Delta \pi^2]_0 \gets 0$
    \For{$t = 1$ to $n$}
        \State $g_t \gets  \ \Pi$
        \State $E[g^2]_t \gets \rho E[g^2]_{t-1} + (1 - \rho) g_t^2$
        \State $\Delta \pi_t \gets - \frac{\sqrt{E[\Delta \pi^2]_{t-1} + \epsilon}}{\sqrt{E[g^2]_t + \epsilon}} g_t$
        \State $E[\Delta \pi^2]_t \gets \rho E[\Delta \pi^2]_{t-1} + (1 - \rho) \Delta \pi_t^2$
        \State $\pi_t \gets \pi_{t-1} + \Delta \pi_t$
    \EndFor
\end{algorithmic}
\end{algorithm}
\subsection{Adaptive Moment Estimation (Adam)}
Adam combines the benefits of RMSProp and momentum by computing adaptive learning rates and moment estimates:
\begin{align}
    m_t &= \bar{\beta}_1 m_{t-1} + (1 - \bar{\beta}_1)  \ \Pi, \\
    v_t &= \bar{\beta}_2 v_{t-1} + (1 - \bar{\beta}_2)  \ \Pi^2, \\
    \hat{m}_t &= \frac{m_t}{1 - \bar{\beta}_1^t}, \quad \hat{v}_t = \frac{v_t}{1 - \bar{\beta}_2^t}, \\
    \pi_{t+1} &= \pi_t - \frac{1}{\sqrt{\hat{v}_t + \epsilon}} \hat{m}_t
\end{align}
\begin{algorithm}[H]
\caption{Adam Optimization for \texttt{$\text{GMP}^{3}$}}
\begin{algorithmic}[0]
    \Require Learning rates $\alpha, \beta_1, \beta_2$, decay rates $\bar{\beta}_1$, $\bar{\beta}_2$, small constant $\epsilon$
    \State Initialize policy $\pi_0$, $m_0 \gets 0$, $v_0 \gets 0$
    \For{$t = 1$ to $n$}
        \State $g_t \gets \ \Pi$
        \State $m_t \gets \beta_1 m_{t-1} + (1 - \beta_1) g_t$
        \State $v_t \gets \beta_2 v_{t-1} + (1 - \beta_2) g_t^2$
        \State $\hat{m}_t \gets \frac{m_t}{1 - \beta_1^t}$
        \State $\hat{v}_t \gets \frac{v_t}{1 - \beta_2^t}$
        \State $\pi_t \gets \pi_{t-1} - \frac{\eta}{\sqrt{\hat{v}_t + \epsilon}} \hat{m}_t$
    \EndFor
\end{algorithmic}
\end{algorithm}
\begin{remark}
Each of these optimization techniques refines the policy update in the {\texttt{$\text{GMP}^{3}$}} framework, enhancing convergence and stability in reinforcement learning-based trajectory planning.    
\end{remark}
\section{Simulation Results}\label{sec:5}
To evaluate the performance of the proposed  {\texttt{$\text{GMP}^{3}$}} algorithm in a fully $\mathrm{SE}(3)$-aware, we conducted simulations in a 3D environment. The UAV is initialized in pose $\mathbf{T}_s = (0, 0, 0, 0, 0, 0) \in \mathrm{SE}(3)$ and assigned to reach final pose $\mathbf{T}_g = (10, 10, 0, 0, 0, 0)$, where the final orientation is level. The trajectory is parameterized by three internal breakpoints per degree of freedom, yielding a 6$n$-dimensional decision vector $\Theta \in \mathbb{R}^{18}$ that controls the full-body motion (position and orientation) over time.  The environment includes four static spherical obstacles defined by
\begin{align*}
\mathcal{O} = \Big\{ 
&(\mathbf{o}_1, r^1_{\mathrm{obs}}) = ((5,\ 5,\ -4.5),\ 5.0), \\
&(\mathbf{o}_2, r^2_{\mathrm{obs}}) = ((2.1,\ 2.0,\ 1.5),\ 1.0), \\
&(\mathbf{o}_3, r^3_{\mathrm{obs}}) = ((8.0,\ 8.0,\ 1.0),\ 1.0), \\
&(\mathbf{o}_4, r^4_{\mathrm{obs}}) = ((2.0,\ 6.0,\ 1.0),\ 1.0)
\Big\}.
\end{align*}
Obstacle violation is penalized using the cost function described in~\eqref{violation}, integrated into the loss aware of $\mathrm{SE}(3)$ $\mathcal{L}_{\mathrm{SE(3)}}$ that also incorporates translational and rotational smoothness. The obstacle proximity weight $\lambda = 2.5$, and the rotational regularization weight is set to $\mu = 0.1$. The translational cost is evaluated with the positive definite weight matrix $Q = \text{diag}(0.89,\ 0.89,\ 0.89)$.
\begin{figure}[ht]
\centering
    \includegraphics[width=\linewidth]{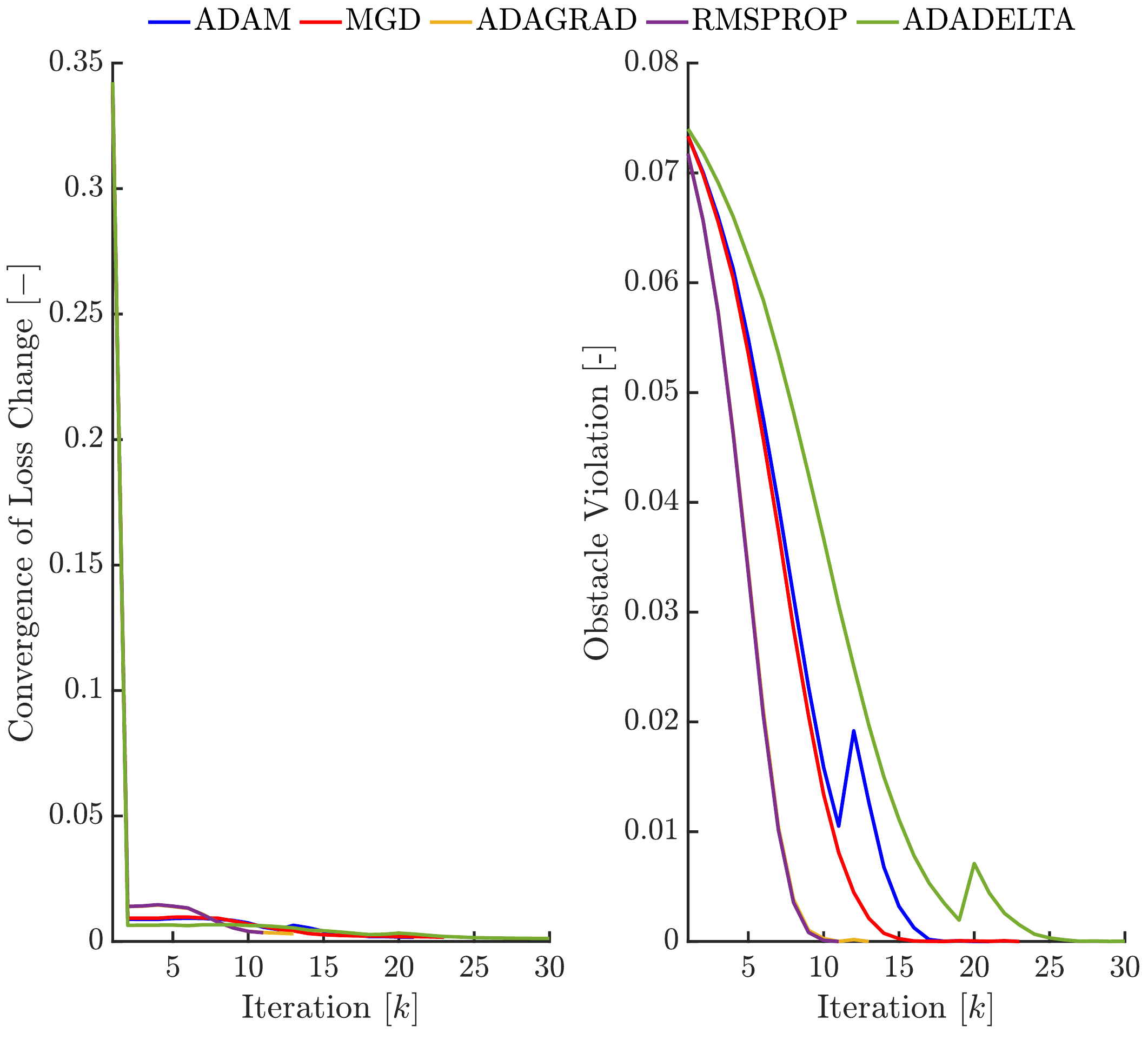}
    \caption{Without influence-aware update}
    \label{fig:comparison_noinf}
    \includegraphics[width=\linewidth]{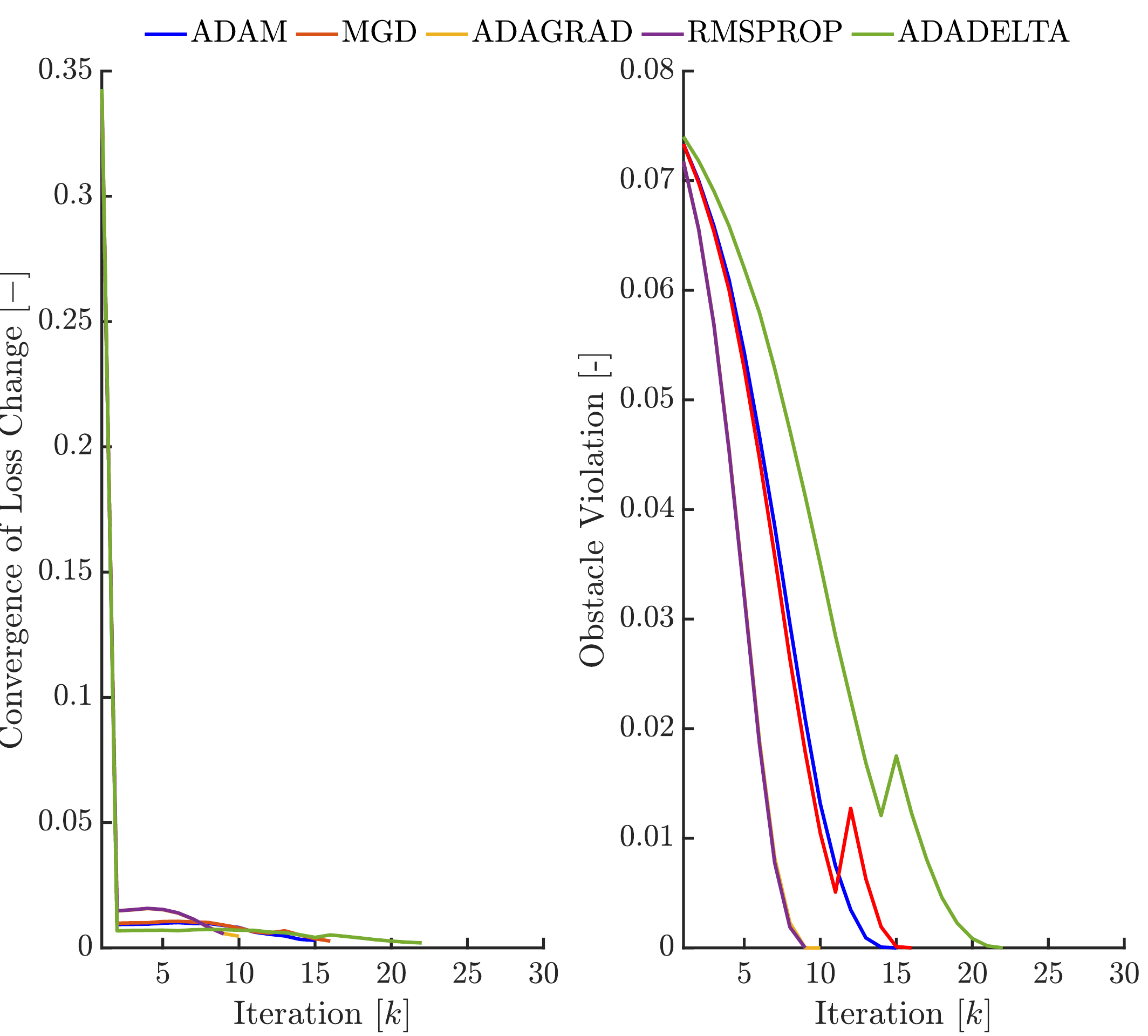}
    \caption{With influence-aware update}
    \label{fig:comparison_inf}
\caption{Comparison of different learning strategies in {\texttt{$\text{GMP}^{3}$}}}
\label{fig:comparison_loss}
\end{figure}
The $\mathrm{SE}(3)$ trajectory is shaped using five learning rules, each executed over $30$ iterations with the same initialization and under the same kinematic and dynamic constraints. The full simulation is conducted using the discrete-time dynamics on $\mathrm{SE}(3)$ in equation~\eqref{SEt}, with time resolution $\Delta t = 0.05\ \mathrm{s}$. Fig.~\ref{fig:3D_ex} shows the learned UAV trajectories for each method, visualized in a three-dimensional space with vertical altitude stalks and translucent spherical obstacles. All methods yield dynamically feasible solutions under bounded velocity constraints. However, distinct differences are observed in the quality of the path and the proximity of the obstacles. Let $\mathbf{R}_{t_j}$ denote the rotation matrix in time $t_j$ derived from the Euler angles $\boldsymbol{\phi}_{t_j} = (\psi_{t_j}, \theta_{t_j}, \phi_{t_j})^\top$. The rotational increment $d_R(\mathbf{R}_{t_j}, \mathbf{R}_{t_{j+1}})$ is computed using the Frobenius norm of the matrix logarithm, as defined earlier. As shown in the bottom row of Fig.~\ref{fig:3D_ex}, both Adam and RMSProp produce smooth and bounded rotational profiles. The angle of yaw $\psi$ remains within $[-\pi,\pi]$ and evolves continuously over the horizon, while the pitch $\theta$ and the roll $\phi$ remain within safe limits as required by the dynamic feasibility of the UAV. In contrast, the MGD and AdaGrad optimizers exhibit higher angular rates and larger oscillations, particularly in pitch, leading to more abrupt orientation changes.

Convergence characteristics for each optimizer are shown in Fig.~\ref{fig:comparison_loss}, which shows the normalized loss difference and the cumulative obstacle violation as a function of iteration $k$.

The RMSProp optimizer achieves the fastest convergence, with a monotonic decrease in loss and minimal final violation. It consistently generates smoother translational and rotational profiles while avoiding obstacles with greater margin. The Adam optimizer exhibits similar convergence speed but slightly higher final loss. The AdaDelta method converges more slowly and yields trajectories with moderate deviation from optimality. Both MGD and AdaGrad show limited convergence and larger obstacle violation, indicating reduced effectiveness under the $\mathrm{SE}(3)$ dynamics and consensus-influenced gradient updates.

Further experiments incorporating influence-aware policy updates—based on the structure of Eq.~\eqref{PU}—are shown in Fig.~\ref{fig:comparison_loss}. The incorporation of the influence-aware policy update, which blends both the best global and local policies, further stabilizes convergence and improves trajectory feasibility in obstacle-rich environments.
\section{DroneManager}\label{sec:6}
\begin{figure*}[htb]
    \centering
    \includegraphics[width=\linewidth]{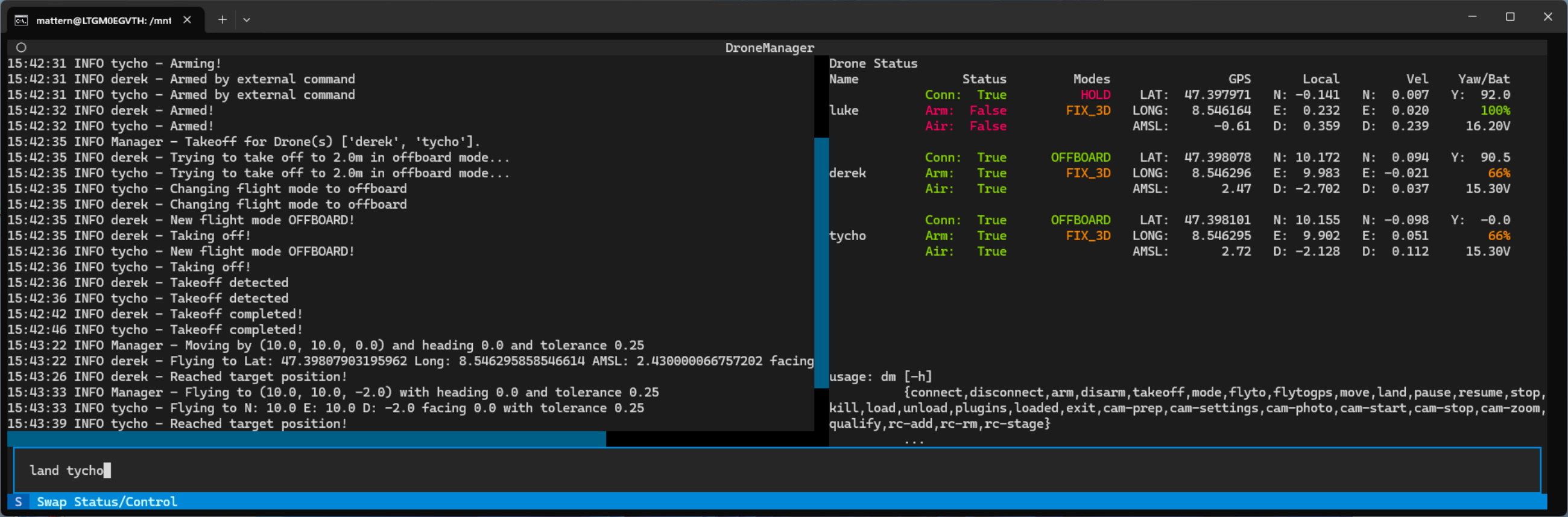}
    \caption{A screenshot of the tool in use. The drones "Derek" and "Tycho" were armed and told to takeoff at the same time, before being sent to different coordinates.}
    \label{fig:toolscreen}
\end{figure*}

We developed a new tool, called DroneManager, to assist in our experiments, as we found existing tools, such as QGroundControl and Mission Planner, difficult to use for multiple drones at once.
It uses MAVLink as the communication protocol, is written in Python and provides functions to connect to and give MAVLink commands, such as takeoff, moving and landing, to any number of drones at once, including queueing commands. It comes with a plugin system to extend the functionality for whatever purposes the user requires, as well as a terminal-based interface that can be used to control drones directly. A screenshot of the interface is shown in Fig.~\ref{fig:toolscreen}.
It works with any platform also using MAVLink, whether simulated or real. We have used it with multiple PX4 and ArduPilot drones, as well as the PX4 Gazebo SITL simulator. DroneManager supports all the basic commands required for flight, such as arming, take-off and landing. Drones are moved with move-by or fly-to commands, either through the text interface or programmatically. These commands support both GPS and local coordinate systems. For indoor use, we also implemented a geo-fence like system that works in the drones local coordinate system and rejects any destinations or setpoints outside the defined flight area.
\subsection{Movement program flow}
For the most part, DroneManager commands only do a little processing to determine the appropriate MAVLink message and then send that to the drone. Movement commands have a more complicated flow with split trajectory control. The command provides a final target position, which is sent to the {\texttt{$\text{GMP}^{3}$}} algorithm, which produces a series of way points, from the current position of the drone to the destination. A trajectory follower algorithm then queries these way points and determines the actual setpoints that will be sent to the flight controller on the drone.

The main purpose of splitting trajectory control into two separate sections is to allow for path planning at different scales. In this concept, the trajectory generator produces a long-term path, accounting for any known obstacles along the way, which the follower algorithm pursues while accounting for any inaccuracies or previously unknown or dynamic obstacles that may appear. For example, imagine an implementation with {\texttt{$\text{GMP}^{3}$}} as a trajectory generator and a potential-based obstacle avoidance algorithm as a trajectory follower. When DroneManager receives a command to fly a drone to some target position, either through the CLI or programmatically, that target position would be sent to {\texttt{$\text{GMP}^{3}$}} to produce a short-distance waypoint, which would then be sent to the avoidance algorithm, that tries to move to the current waypoint while keeping its distance to any previously unknown obstacles. Once the follower algorithm reaches the waypoint, it asks for the next one from the generator, until the final destination is reached. 

The interface requirements for the generator and follower algorithms are kept minimal, allowing a lot of flexibility in their usage. We provide a basic "pass-through" implementation for both, with the generator continually producing the destination position as way points, while the follower simply passes these way points as setpoints directly to the drone. For our experiments, we implemented \texttt{$\text{GMP}^{3}$} as a trajectory generator, which selects the way point on the trajectory produced by \texttt{$\text{GMP}^{3}$} that matches the current timestamp, and used the pass-through implementation of a trajectory follower to send these way points directly to the drone. \texttt{$\text{GMP}^{3}$} is executed in a separate process to ensure that critical functions are not blocked.

\subsection{Plugin system}
DroneManager also features a simple plugin system for adding new features. Plugins can be loaded and unloaded either with a text command or programmatically and provide extra features. The purpose of this modular approach is to keep runtime and memory requirements minimal, and to reduce clutter in the interface. A plugin is a python module located in the plugins folder with exactly one class whose name ends with "Plugin" and that subclasses an abstract Plugin base class. This base class handles the basic loading and unloading process and provides attributes to simplify working with the rest of the software. These include other plugin dependencies, a list of background coroutines that are started and stopped automatically on loading and unloading, and a list of CLI functions, for which new text commands, including type checking and help texts, are automatically generated for the UI by inspecting the signature of the corresponding functions when the plugin is loaded. Plugins are added as attributes to the DroneManager instance under their own name, allowing easy programmatic access to them from other parts of the software. A number of current features, such as the mission system, are implemented as plugins.
\subsection{Mission system}
Missions are implemented as a type of plugin and use a lot of the same systems. Unlike basic plugins, they can be given custom names, which must be unique, to allow multiple missions of the same type to run simultaneously. As with the trajectory system, the interface requirements are kept minimal. Missions must contain drones, with functions to add and remove drones from missions, take place in some kind of flight area and be able to provide some kind of status message to the user. Other than that, implementations are free to be as complex or simple as required.

We have used this system during multiple demonstration flights where up to four drones were tasked with scanning a field in formation and locating a dummy lying on the ground, including maintaining observation of the dummy by continually swapping the observing drone with stand-by drones once its battery started running low. The mission flow was broken up into multiple stages, with stage transitions triggering automatically once the previous stage had completed, except for the commands to start and stop the mission. The mission started in a "Ready" stage, a text command then launched the "Search" stage, wherein the drones scanned the search area until either the whole area was scanned or the dummy was found. If the dummy was found a transition stage sent all searching drones except one back to their launch points, and then moved to the long-term observation stage, wherein the observing drone slowly circled the dummy. Once the observing drones battery ran low, the stand-by drone with the highest battery was launched and a swap took place, with the old observing drone flying back to its launch point to have its battery replaced. This swapping took place indefinitely until another text command was sent to return to base, at which point all drones in flight returned to their launch points and the mission concluded.

\begin{figure}[h]
    \centering
    \includegraphics[width=\linewidth]{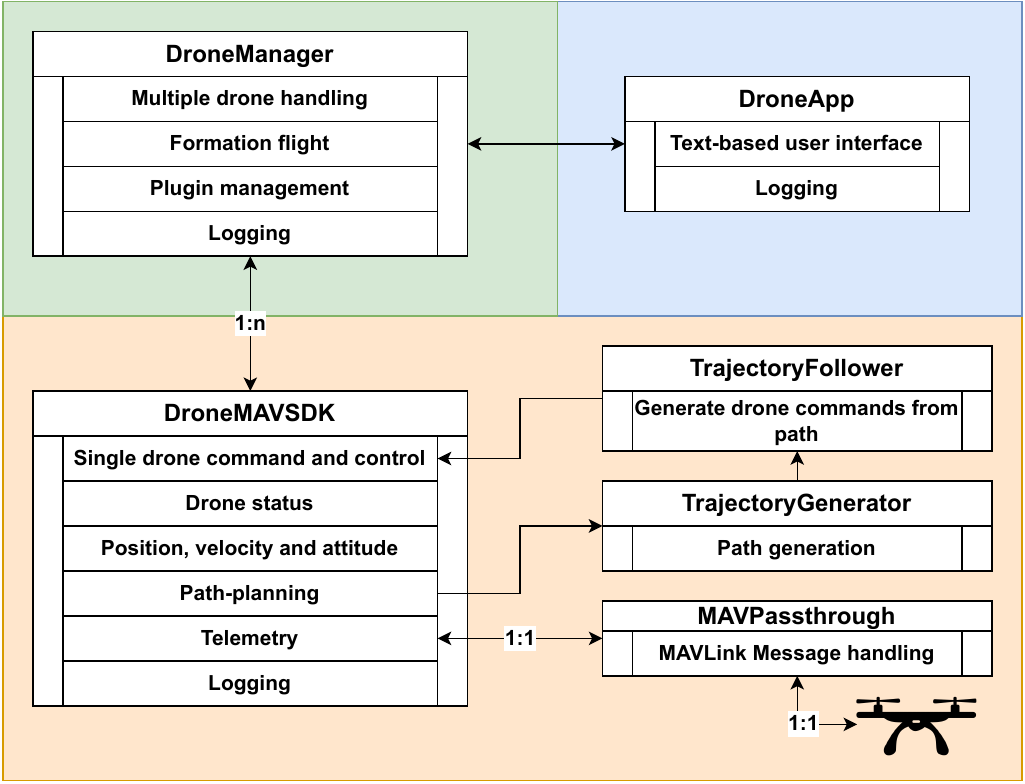}
    \caption{A chart showing the different components of the software. The green area is the core of the program, the blue section deals with user interfaces and the orange section handles individual drones.}
    \label{fig:impchart}
\end{figure}

A webpage based visualization was used to show the state of the mission to viewers, with information sent from DroneManager to the web server using UDP, with this communication system implemented as another plugin. The web-based visualization is not currently part of DroneManager. A schematic with the different components and their purposes is shown in Fig.~\ref{fig:impchart}.
\section{Practical Experiment}\label{sec:7}
\begin{figure}[h]
    \centering
    \includegraphics[width=\linewidth]{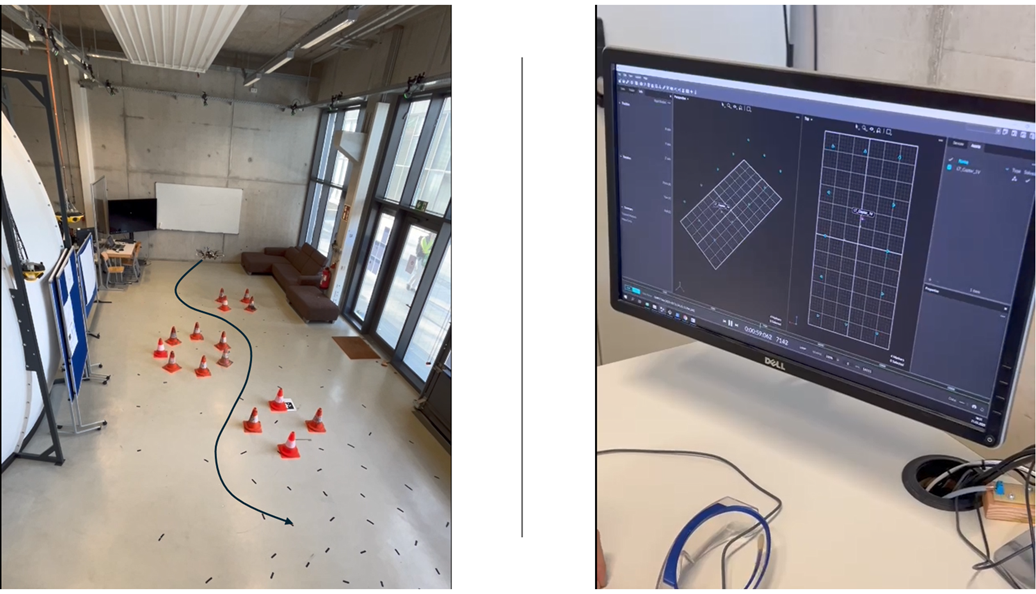}
    \caption{Experimental setup used for evaluating the {\texttt{$\text{GMP}^{3}$}} algorithm in a real-world scenario. The drone is tasked with navigating a predefined path in an indoor environment while avoiding marked obstacles (indicated by cones). The setup includes a motion tracking system for precise localization and a GCS executing the {\texttt{$\text{GMP}^{3}$}} trajectory planner in real time.}
    \label{experimental}
\end{figure}
\begin{figure}[h]
    \centering
    \includegraphics[width=\linewidth]{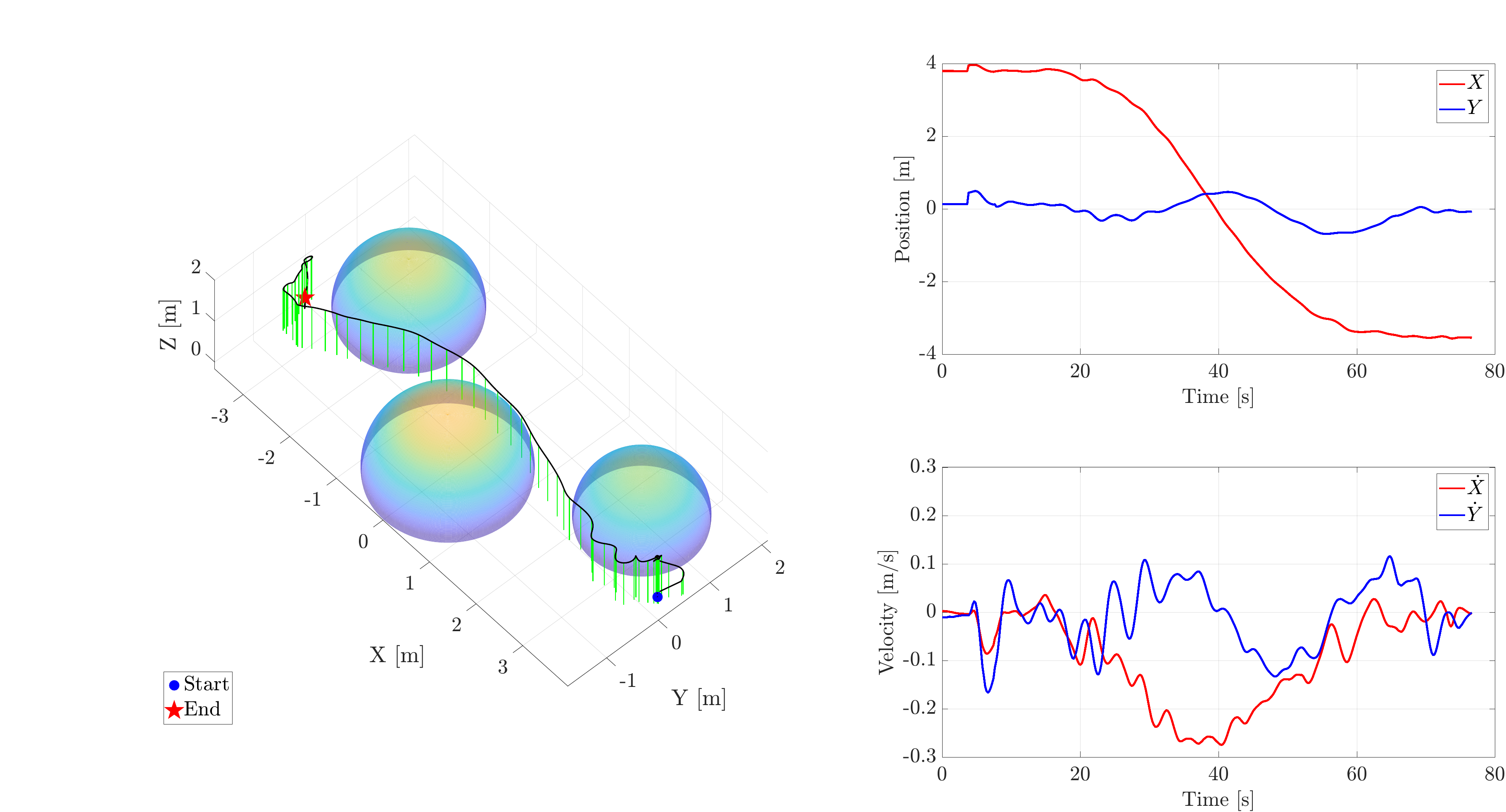}
    \caption{Extracted onboard flight data from the experiment \textit{Left:} Obstacles and Trajectory of the UAV. \textit{Right:} Position and Velocities of the UAV over its flight time.}
    \label{fig:exp_flight}
\end{figure}

In this section, we present an experiment designed to (i) demonstrate the effectiveness of the \texttt{$\text{GMP}^{3}$} algorithm and (ii) highlight its advantages in practical applications. The experimental setup is depicted in Fig.\ref{experimental}.
In the demonstration, a drone is tasked with flying from one point to another while avoiding obstacles, marked as circles on the ground using traffic cones. Due to regulatory constraints, the demonstration took place indoors using a near-infrared motion tracking system for positioning. The control and coordination of the drone were handled through our custom-developed software, DroneManager, with the {\texttt{$\text{GMP}^{3}$}} algorithm integrated as the trajectory generation module. As described in the previous section, DroneManager allowed real-time policy updates, trajectory execution, and safety constraints enforcement in a modular and extensible environment. A graphic of the flight area with obstacles, the planned trajectory and the actual trajectory is shown in Fig. \ref{fig:exp_flight}. 

The location of the obstacles was known beforehand, no object detection was taking place. For safety, we limited the drones speed to about 25 cm per second and configured our GCS to reject any set points outside the flight area. The tracking system and GCS operated independently. The drones position was updated by the tracking system at 120Hz while the drone sent position and attitude information to the GCS at 5Hz. \texttt{$\text{GMP}^{3}$} was also configured with a temporal resolution of 5Hz. Other \texttt{$\text{GMP}^{3}$} parameters are available in appendix~\ref{app:gmpparams}.
\section{CONCLUSION}
We presented {\texttt{$\text{GMP}^{3}$}}, a reinforcement learning-based framework for global multi-phase trajectory planning that operates directly on the Lie group $\mathrm{SE}(3)$, enabling full 6-DoF motion planning for UAVs in cluttered environments. By formulating trajectory optimization as a distributed learning problem over rigid-body poses, and introducing a Bellman operator tailored to consensus-aware $\mathrm{SE}(3)$ dynamics, the framework enables cooperative refinement of both translational and rotational motion under physical constraints.

The integration of {\texttt{$\text{GMP}^{3}$}} with the custom-developed DroneManager software allows real-time deployment on real-world UAV platforms, supporting modular control, trajectory tracking, and safety enforcement via MAVLink-based communication. Extensive simulations and practical experiments confirm the framework’s ability to generate smooth, feasible, and collision-free trajectories while respecting kinematic limits and rotational consistency. Among various optimizers evaluated, RMSProp consistently demonstrated the most favorable convergence behavior and violation minimization in the $\mathrm{SE}(3)$ setting.

These findings underline the utility of {\texttt{$\text{GMP}^{3}$}} as a scalable and reliable trajectory planning solution for UAVs operating in dynamic or safety-critical environments. Future work will explore the extension of {\texttt{$\text{GMP}^{3}$}} to coordinated multi-agent scenarios in $\mathrm{SE}(3)$, leveraging DroneManager’s infrastructure for real-time formation control, decentralized communication, and adaptive leader-follower strategies in partially known or rapidly evolving workspaces.

\appendices
\section{\texttt{$\text{GMP}^{3}$} Parameters (RMSProp)}
\label{app:gmpparams}
\begin{table}[h]
    \centering
    \begin{tabular}{c|cc}
        Parameter & Value & Description\\
        \hline
        Max. Iterations & 100 & Max. number of iterations before failure\\
        $\gamma$ & 0.01 & {Decay factor}\\
        Q$_{11}$ & 0.89 & {Weight for Dimension X}\\
        Q$_{22}$ & 0.89 & {Weight for Dimension Y}\\
        Q$_{33}$ & 0.89 & {Weight for Dimension Z}\\
        $\alpha$ & 0.0028 & {Learning rate} \\
        $\beta_1$ & 0.0028 & {-} \\
        $\beta_2$ & 0.0028 & {-} \\
        $dt$ & 0.2 & Time-resolution of trajectory\\
    \end{tabular}
\end{table}


\bibliographystyle{IEEEtran}
\bibliography{IEEEabrv,references}

\end{document}